\documentclass[preprint,12pt]{elsarticle}

\usepackage{epsfig}
\usepackage{graphicx}
\usepackage{amsmath}
\usepackage{amssymb}
\usepackage{mathtools}
\usepackage{dsfont}
\usepackage{amsthm}
\DeclareMathOperator*{\argmin}{arg\,min}
\usepackage{multirow} 
\usepackage[table]{xcolor}
\usepackage{pgfplots}
\pgfplotsset{compat=1.9}
\usetikzlibrary{positioning}
\usepackage{tikz}
\usetikzlibrary{positioning}
\usetikzlibrary{shapes.geometric}

\usepackage{booktabs}
\usepackage{array}
\newcommand\Tstrut{\rule{0pt}{3mm}}
\newcommand\Tstrutt{\rule{0pt}{5mm}}
\usepackage{adjustbox}
\usepgfplotslibrary{groupplots}
\usetikzlibrary{matrix}
\usepackage[T1]{fontenc}
\usepackage[hidelinks]{hyperref}

\definecolor{airforceblue}{rgb}{0.36, 0.54, 0.66}
\definecolor{amber}{rgb}{1.0, 0.49, 0.0}
\definecolor{atomictangerine}{rgb}{1.0, 0.6, 0.4}
\definecolor{ballblue}{rgb}{0.13, 0.67, 0.8}
\definecolor{babyblueeyes}{rgb}{0.63, 0.79, 0.95}
\definecolor{babyblue}{rgb}{0.54, 0.81, 0.94}
\definecolor{applegreen}{rgb}{0.55, 0.71, 0.0}
\definecolor{ashgrey}{rgb}{0.7, 0.75, 0.71}
\definecolor{aurometalsaurus}{rgb}{0.43, 0.5, 0.5}

\journal{Signal Processing}
\begin{document}
\begin{frontmatter}



\title{Aggregated \texorpdfstring{$f$}{f}-average Neural Network applied to Few-Shot Class Incremental Learning}

\author[cvn]{Mathieu Vu \corref{cor1}\fnref{fnref1}}
\author[cvn]{Émilie Chouzenoux\fnref{fnref1}}
\author[ets]{Ismail Ben Ayed}
\author[cvn]{Jean-Christophe Pesquet}

\fntext[fnref1]{M. Vu and E. Chouzenoux acknowledge support from the European Research Council under Starting Grant MAJORIS ERC-2019-STG-850925.}
\cortext[cor1]{mathieu.vu@inria.fr}

\affiliation[cvn]{organization={OPIS - CVN, Inria Saclay, CentraleSupélec, Université Paris Saclay},
            addressline={3 rue Joliot Curie},
            city={Gif-sur-Yvette},
            postcode={91190},
            state={Île-de-France},
            country={France}}

\affiliation[ets]{organization={LIVIA - École de Technologie Supérieure},
            addressline={1100 rue Notre Dame Ouest},
            city={Montréal},
            postcode={H3C 1K3},
            state={Québec},
            country={Canada}}

\begin{abstract}
Ensemble learning leverages multiple models (i.e., weak learners) on a common machine learning task to enhance prediction performance. Basic ensembling approaches average weak learners outputs, while more sophisticated ones stack a machine learning model in between the weak learners outputs and the final prediction. This work merges both aforementioned frameworks. We introduce an \emph{aggregated $f$-averages} (AFA) shallow neural network which models and combines different types of averages to perform an optimal aggregation of the weak learners predictions. We emphasise its interpretable architecture and simple training strategy and illustrate its good performance on the problem of few-shot class incremental learning.
\end{abstract}

\begin{graphicalabstract}
\begin{figure}[ht!]
    \centering
    \begin{tikzpicture}[
            snode/.style={rectangle, draw, minimum width=1cm, minimum height=1cm, text=white},
            dnode/.style={cylinder, draw, aspect = 0.2, shape border rotate = 90, text=white},
            scale=0.85, every node/.style={transform shape}
        ]

        \node (s0) at (-11,6)  {Session};
        \node (s1) at (-11,5)  {1};
        \node (s2) at (-11,1)  {2};
        \node (s3) at (-11,-2.5) {$\vdots$};
        \node (s4) at (-11,-5) {$K$};
        \node[dnode, fill=amber] (d1) at (-9.5,5)  {$D_1^\text{train}$};
        \node[dnode, fill=amber] (d2) at (-9.5,1)  {$D_2^\text{train}$};
        \node[]      (d3) at (-9.5,-2.5) {$\vdots$};
        \node[dnode, fill=amber] (d4) at (-9.5,-5) {$D_K^\text{train}$};
        \node[snode, fill=airforceblue] (m1) at (-7.5,5)  {$f_\theta$}; 
        \node[snode, fill=babyblue] (m1h) [right=0cm of m1, align=center]  {$H_{W_1}$};
        \node[snode, fill=airforceblue] (m2) at (-7.5,1)  {$f_\theta$};
        \node[snode, fill=babyblueeyes] (m2h) [right=0cm of m2, align=center]  {$H_{W_2}$};
        \node[]      (m3) at (-7,-2.5) {$\vdots$};
        \node[snode, fill=airforceblue] (m4) at (-7.5,-5) {$f_\theta$};
        \node[snode, fill=babyblueeyes] (m4h) [right=0cm of m4, align=center]  {$H_{W_K}$};

        \node[] (o1) at (-4,5) {
            $\begin{pmatrix}
                x_{1,1} \\ 
                \vdots \\ 
                x_{1, N_1} \\
                \textcolor{aurometalsaurus}{p_{1,1}} \\
                \textcolor{aurometalsaurus}{\vdots} \\
                \textcolor{aurometalsaurus}{p_{1, N_K - N_1}} \\
            \end{pmatrix}$
        };
        \node[] (o2) at (-4,1) {
            $\begin{pmatrix}
                x_{2,1} \\ 
                \vdots \\ 
                x_{2, N_2} \\
                \textcolor{aurometalsaurus}{p_{2,1}} \\
                \textcolor{aurometalsaurus}{\vdots} \\
                \textcolor{aurometalsaurus}{p_{2, N_K - N_2}} \\
            \end{pmatrix}$
        };
        \node[] (o3) at (-4,-2.5) {\rotatebox{90}{$\cdots$}};
        \node[] (o4) at (-4,-5) {
            $\begin{pmatrix}
                x_{K,1} \\ 
                x_{K,2} \\ 
                \vdots \\ 
                x_{K, N_K} \\
            \end{pmatrix}$
        };
        \node[snode, minimum height=12cm, minimum width=3cm, align=center, fill=applegreen] (e) at (0,0) {Aggregated \\ $f$-averages};
        \node (x) at (3,0) {
            $\begin{pmatrix}
                \hat{x}_{1} \\
                \vdots            \\
                \hat{x}_{N_K}
            \end{pmatrix}$
        };

        \draw[-latex] (d1.east) -- (m1.west);
        \draw[-latex] (d2.east) -- (m2.west);
        \draw[-latex] (d4.east) -- (m4.west);
        \draw[-latex] (m1h.east) -- (o1.west);
        \draw[-latex] (m2h.east) -- (o2.west);
        \draw[-latex] (m4h.east) -- (o4.west);
        \draw[-latex] (o1.east) -- (o1-|e.west);
        \draw[-latex] (o2.east) -- (o2-|e.west);
        \draw[-latex] (o4.east) -- (o4-|e.west);
        \draw[-latex] (e.east) -- (x.west);
    \end{tikzpicture}
    \caption{This work addresses few-shot class incremental learning by training an ensemble of weak classifiers, each specialised in their own set of classes from their given session. For each session $k$, the weak classifier is composed of a feature extractor $f_\theta$ (trained only in base session and frozen in subsequent sessions) and a classification head $H_{W_k}$ (fitted on data from their own new session $k$). A dynamic padding compensates the discrepancy between the sizes of prediction vectors at each session. The resulting set of weak classifiers is ensembled by our proposed Aggregated $f$-averages (AFA) model. AFA neural network is designed to model and combine different types of averages, in an automatic supervised fashion, in order to perform an optimal output fusion.}
\end{figure}
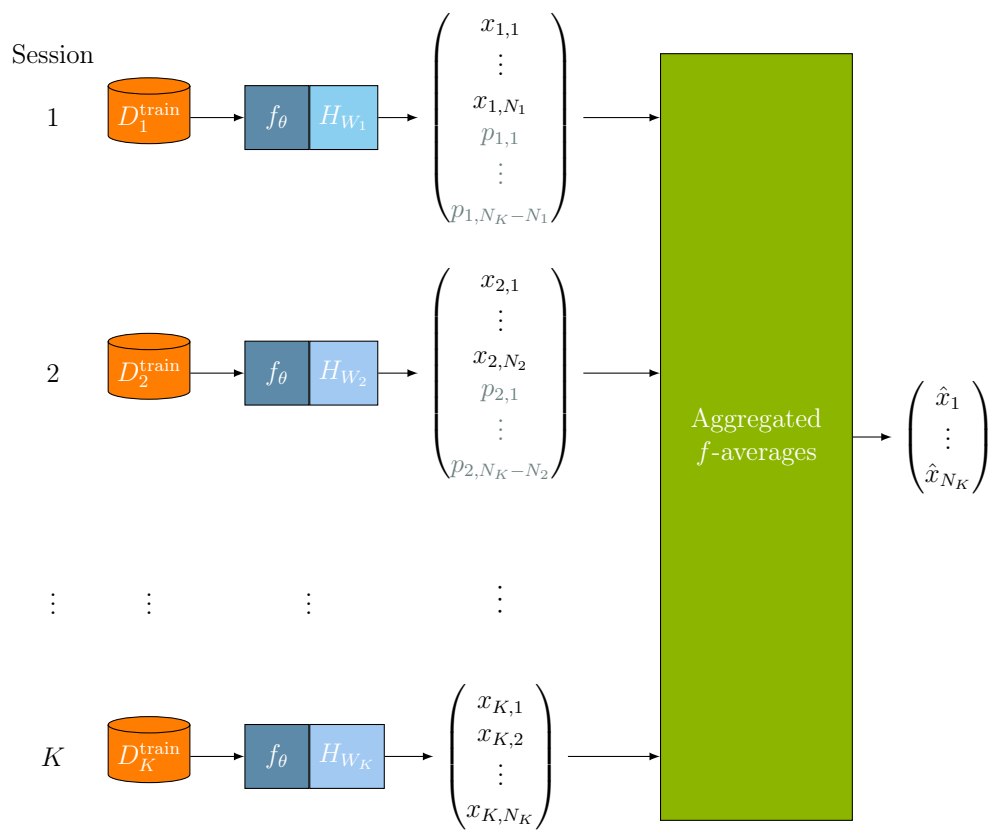
\end{graphicalabstract}

\begin{highlights}
\item We address the problem of few shot class incremental learning (FSCIL) as a combination of successive few shot learning problems, by relying on the ensemble learning paradigm. We ensemble a set of weak classifiers, each specialised in their own session and their own set of classes.
\item For an efficient ensembling, we introduce a novel method called \emph{Aggregated $f$-averages} (AFA). Our ensembling method is a supervised neural network model with a specific architecture, specific activation functions, and specific weight constraints, enabling to model and combine, in an automatic supervised manner, various types of averages.
\item We show that our proposed method sensibly outperforms classic output fusion approaches for ensemble learning and yields comparable results with respect to models specifically dedicated to FSCIL. 
\end{highlights}

\end{frontmatter}


\section{Introduction}
Deep Convolutional Neural Networks (CNNs) have become the state-of-the-art reference in most computer vision problems. They rely on curated datasets, for which quality and quantity have a critical impact on the model performances. For instance, a challenging scenario arises when classifying images belonging to new (unseen) classes, with only a few annotations. More specifically, in supervised classification, models are only able to classify categories that were seen during training phase. If unseen classes are added, it is not straightforward to handle them without retraining the whole model. Incremental learning aims at designing solutions for this situation. However, when upcoming classes include only a few training samples, incremental learning methods do not perform well. Few-Shot Class Incremental Learning (FSCIL), a challenging scenario introduced recently in \cite{tao2020fscil}, focuses on designing methods that can deal with both incremental and few-shot settings. On the one hand, a small training set may induce overfitting, hindering classification performances on the base classes. This phenomenon is known as \emph{catastrophic forgetting}. On the other hand, the focus on avoiding catastrophic forgettingmay have a counterproductive effect as it impedes the learning of novel classes.

To address FSCIL, we propose to consider it as a combination of successive few-shot learning problems. More precisely, we learn a dedicated model for each novel set of class of a new session. These models (called \emph{weak classifiers} in 
ensemble learning) are, in our case, simple nearest-neighbour classifiers based on class prototypes. Since they are specialised on their own set of classes, they may not perform well on classes that were not included in their training set (whether they were seen before or after). This is why we train a final ensembling model, which performs predictions on all classes by integrating the weak classifiers. To perform the ensembling, we introduce a novel method called \emph{Aggregated $f$-averages}. Our ensembling method is a supervised neural network model with a specific architecture, specific activation functions and specific weight constraints, enabling to model and combine various types of averages. It uses a few parameters while being interpretable. This yields a strong FSCIL baseline, which integrates simple prototype-based classifiers and performs competitively in comparisons to specialised and convoluted state-of-the-art FSCIL techniques on several datasets. Moreover, we show experimentally the substantial effect of our $f$-averages ensembling in comparisons to standard ensembling methods. Also, we closely examine how FSCIL methods are evaluated, and argue that the average accuracy over all classes may not be appropriate in FSCIL as the problem is an instance of imbalanced classification. For instance, a model trained only on the base classes, while ignoring subsequent new classes, has a rather steady mean accuracy over all the classes when the number of base classes is greater than the total number of new classes. Instead, we advocate the use of the F1 score, which is commonly used in imbalanced classification problems.

\section{Related Work}

\subsection{The standard few-shot classification setting}
Deep learning models have reached human-level performance in many tasks thanks to large scale datasets. However, their generalisation abilities are challenged in settings where only a few training samples are available for new, unseen classes. Numerous research articles have attempted to tackle this few-shot classification problem, including for example \cite{can,chen2019closer,simpleshot,kim2019edge,relation_net,martin2022towards,boudiafadvances,prototypical_nets,maml}. The few-shot classification setting consists of two stages: first, in the base stage, a model is trained with a substantial amount of labelled data on a set of base classes; then comes the generalisation stage where only a small number of labelled training samples are
available for the novel classes.

In the standard few-shot setting, the literature abounds of convoluted methods based on meta-learning, which consists of "learning to learn" \cite{maml,prototypical_nets,matching_net,relation_net,lee2019meta}. The base stage is used to perform a series of \emph{episodic training} that will simulate the generalisation stage. For example, prototypical networks \cite{prototypical_nets} meta-learns how to represent each class with an embedding prototype and to maximise the log-probability of test samples. Similarly, MAML \cite{maml} is a method producing a meta-learner that helps the model to better fine-tune on novel classes during generalisation stage. However, recent studies have shown that standard training strategies coupled with transfer learning outperforms most meta-learning methods \cite{chen2019closer, boudiafadvances, simpleshot}. For instance, SimpleShot \cite{simpleshot} performs a basic nearest-neighbour classifier based on class prototypes, outperforming most meta-learning methods. 

\subsection{Few-shot Class Incremental Learning}

Few-shot Class Incremental Learning (FSCIL), recently introduced in \cite{tao2020fscil}, attempts to address the poor performances of CIL methods \cite{castro2018end, hou2019learning} when only a few training samples are available for the new classes. 
A variety of specialised and, in most cases, convoluted methods (e.g., based on meta-learning) were recently developed for FSCIL.  
For instance, TOPIC \cite{tao2020fscil} originally implements a neural gas network, which preserves the topology of the feature space when learning new classes, 
thereby mitigating the forgetting of previously learnt classes. IDVLQ-C's \cite{chen2020incremental} classification head is built using reference vectors that are prototypes optimised through a custom loss, avoiding overlap. State-of-the-art methods achieve best results using even more elaborate classification heads combined with meta-learning-inspired training schemes. For example, SPPR \cite{zhu2021self} generates incremental episode during each iteration of base training. Furthermore, new-class samples are leveraged in a self-promoted prototype refinement mechanism that compute their relations with respect to previous training samples in a latent space. Similarly, CEC \cite{zhang2021few} meta-learns a Graph Attention Network (GAT) \cite{VelickovicCCRLB18} in a pseudo-incremental learning stage. This external module is able to adapt new classification heads 
given context from previous sessions and new-class samples. C-FSCIL \cite{hersche2022constrained} also meta-learns a module called Explicit Memory (EM) and embedding networks that form a content-based attention mechanism. This takes advantage of prototypes stored in the EM similarly to episodic memory introduced by CIL methods.

The work in \cite{kukleva2021generalized} argued against relying solely on the average accuracy over all the classes as evaluation metric, due to the class-imbalanced nature of the FSCIL problem. This issue
is further exacerbated in earlier sessions, with only a few new classes compared to the many classes of the base sessions. Instead, two metrics are used: average accuracy on base classes and average accuracy on new classes. Using these two metrics enable to disentangle the discovery of new classes and the prevention of catastrophic forgetting.

\subsection{Ensemble learning}
Ensemble learning is a set of methods which leverage an ensemble of models (also called weak learners) instead of relying on a single learner to perform a certain task. While ensembling is obviously more demanding in terms of computing power, it can achieve better accuracy and generalisation, improve overall stability and reduce prediction variance and bias. Two main phases are identified in the process of building an ensemble model: weak learners training and output fusion \cite{sagi2018ensemblesurvey}. The former focuses on producing an ensemble of diverse models, which is a crucial principle in ensemble learning. For example, bootstrap aggregating (also known as bagging) \cite{breiman1996bagging} is a method that trains each model on a different subset of the training data to produce diverse weak learners. The latter, output fusion, consists of gathering outputs from every weak learner of the ensemble and on combining them in order to produce the final prediction.

Two categories of methods can be distinguished to perform output fusion in ensemble learning \cite{sagi2018ensemblesurvey}. The most direct and naive method to produce the unified prediction is to average weak learners outputs or, in the case of classification, to use a majority voting scheme. Different types of averages could be used (e.g., arithmetic, geometric, harmonic, etc), or weights that could be included to further refine results. Those weights can be set using various kinds of criteria like, for example, based on weak learners isolated performances \cite{freund1997decision}. The second category makes use of meta-learners that are an additional model, which is responsible for taking advantage of the weak learners. Mixture of experts is a variant of meta-learners, where a gating network does not combine weak learners output but rather select the weak learner that is most suited to produce the correct prediction given a certain input \cite{jacobs1991adaptive}. A more straightforward output fusion based on meta-learners is the stacking of the additional learning model. Taking weak learners output as input, it learns the best combination to assemble the unified prediction \cite{wolpert1992stacked}. 

\subsection{Our contributions}

As mentioned earlier, state-of-the-art FSCIL methods follow the methodological trend that have dominated the standard few-shot setting by relying on complex meta-learning strategies. We propose here a simpler approach, where the ensemble learning paradigm is used, with the aim to optimally fuse a set 
of nearest-neighbour classifiers based on class prototypes. Given the configuration of the FSCIL setting, training data is characteristically split into subsets whose classes are exclusive. Using those subsets to train weak learners 
naturally provides the desired diversity. Furthermore, we introduce an original ensembling method, which focuses on the output-fusion part of ensemble learning. We show that, in the FSCIL setting, our ensembling model significantly outperforms the output-fusion strategies commonly used in the ensemble learning literature, while being competitive with state-of-the-art FSCIL methods.

Our ensembling method enables to model various types of generalised weighted averages, and to combine and/or select them optimally, in a supervised manner. Moreover, our specific architecture models explicitly the balancing weights between the different types of averages, and hence yields an interpretable ensembling mechanism.
In order to train our ensemble learning model in the context of FSCIL, we adopt an episodic memory to learn the balance between base and new classes. Initial training images are not stored. Instead, only their projection in the feature space is needed, which is memory efficient.

Our experimental analysis relies on advanced metrics (average accuracy on base classes, and average accuracy on new classes) to better understand and assess the trade-off between base and new classes, for the benchmarked models. We also use those metrics to illustrate that the mean F1 score is better suited to evaluate FSCIL methods, as it is a metric commonly used in imbalanced classification tasks.

\section{Problem set-up}

In this work, we design a novel ensemble learning framework applicable to the problem of few-shot class incremental learning (FSCIL). Let us recall the FSCIL problem set-up, in this Section. FSCIL, recently introduced in \cite{tao2020fscil}, focuses on designing machine learning methods that can deal with both incremental \cite{castro2018end, hou2019learning} and few-shot situations \cite{martin2022towards, boudiafadvances}. Specifically, FSCIL aims at including an increasing number of categories in a classification problem with the extra constraint that only a small number of training samples are available for upcoming classes. This is motivated by the frequent practical situation in computer vision when a model, built to classify certain categories seen during training phase, must be adjusted to classify images belonging to new classes with only (very) few annotations. The main difficulty revolves around the ability to learn new classes while preventing \emph{catastrophic forgetting} of classes previously learnt, yielding poor performance of standard incremental learning methods. 

The FSCIL setting considered in this paper consists of successive sessions that incrementally provide new categories of images to be classified. First session, also called base session, is a standard classification problem with a substantial number of training samples for each base class. The number of base classes, that we will denote $n_\text{class\_base}$, is usually large relatively to the total number of classes, that we will denote $n_\text{class}$. In subsequent sessions, only a few training samples are provided for a limited number of novel classes. Reusing standard incremental and few-shot classification notations, let $K$ be the total number of sessions, $n_\text{way}$ the number of new classes and $n_\text{shots}$ the number of samples for each new class. Let $D_k^\text{train}$ (respectively $D_k^\text{test}$) be the sets of training (respectively test) images associated with session $k$ with $k \in \{1,\ldots,K\}$. Similarly, let us denote $C_k^\text{train}$ (respectively $C_k^\text{test}$) the classes labels available at session $k$. During training, for sessions $k \in \{2,\ldots,K\}$, the number of new classes $|C_k^\text{train}|$ equals $n_\text{way}$ and the number of sample images $|D_k^\text{train}|$ equals $ n_\text{shots}$. For base session (i.e., $k=1$), we have a relatively large number of base classes $|C_1^\text{train}| = n_\text{class\_base}$, and a large number of samples $|D_1^\text{train}|$. After each session $k \in \{1,\ldots,K\}$, the goal is to predict, for each image $i \in D_k^\text{test}$, its class label $c$ among the available $C_k^{\text{test}}$. Let us emphasise that $c$ can be any class from base to current session, i.e. that $C_k^\text{test} = \bigcup_{j=1}^k C_j^\text{train}$. At each session $k$, the number of classes to predict $|C_k^\text{test}|$ which we will denote $N_k$ is thus,
\begin{equation} \label{eq:classFSCIL}
    N_k = n_{\text{class\_base}} + (k-1) n_{\text{way}}.
\end{equation} 

\section{Methodology}
\label{sec:methodology}

We now present our main contribution, that is the proposition of a novel ensemble learning approach, and its specification for tackling the problem of FCSIL. We first recall, in Sec.~\ref{sec:ensemAv}, the principle of ensembling based on averaging operations. Then, in Sec.~\ref{sec:ensfa}, we introduce the concept of $f$-average, generalizing the Kolmogorov's mean framework, that will be at the core of our ensemble learning model. In Sec.~\ref{subsection:agg_f_average}, we present our proposed ensembling method called AFA (aggregated $f$-average), and finally, Sec.~\ref{sec:few_shot_ensembling} describes our proposed strategy to integrate ensemble learning in the context of FCSIL. 

\subsection{Ensembling through averaging}
\label{sec:ensemAv}
Consider $K$ machine learning models trained for a common task (e.g., classification), producing $K$ outputs $(x_{k})_{1\le k \le K}$, assumed to be vectors in $\mathbb{R}^N$. In ensemble learning, those $K$ outputs are combined during an output fusion phase in order to produce a single, expectably better, prediction for the task at hand. A naive method is to average the outputs with appropriate weights. We summarize in Table~\ref{tab:weighted_averages} common expressions for weighted averages, with $(\omega_{k})_{1\le k \le K}$ nonnegative reals such that $\sum_{k=1}^{K} \omega_{k} = 1$. 

\bgroup
\def\arraystretch{1.5}
\begin{table*}[!ht]
    \centering
    \caption{Examples of weighted averages along with their validity domains.}
    \begin{adjustbox}{width=\textwidth}
    \begin{tabular}{l|cccc}
        Mean & Arithmetic & Geometric & Harmonic & Power-$q$\\ \hline
        Formula & $\sum_{k=1}^K \omega_{k}x_{k}$ & $\prod_{k=1}^K x_{k}^{\omega_{k}}$ & $\left({\sum_{k=1}^K \frac{\omega_{k}}{x_{k}}}\right)^{-1}$ & $\left( \sum_{k=1}^K \omega_{k} x_{k}^q \right)^{1/q}$  \Tstrutt \\
        Validity & $x_k \in \mathbb{R}^N$ & $x_k \in [0,+\infty)^N$ & $x_k \in (0,+\infty)^N$ & $x_k \in [0,+\infty)^N$, $q>0$  \Tstrutt
    \end{tabular}
    \end{adjustbox}
    \label{tab:weighted_averages}
\end{table*}
\egroup

\subsection{\texorpdfstring{$f$}{f}-average}
\label{sec:ensfa}

Following Kolmogorov's mean framework~\cite{kolmogorov_1930_mathematics}, we can rewrite the above examples under the general form:
\begin{equation}\label{e:genav}
\tilde{x} = f^{-1}\Big(\sum_{k=1}^K \omega_{k} f(x_{k})\Big).
\end{equation}
Hereabove, $f$ is a bijective function from $[0,+\infty)^N$ to some convex $C$ of $\mathbb{R}^N$,
and $f^{-1}$ is its inverse function from $C$ to $[0,+\infty)^N$. We will subsequently assume that 
$f$ operates component-wise, in the sense that it consists of the application of the same scalar function to each of the components of its argument.
Let us now express functions $f$ and $f^{-1}$ to retrieve the popular averaging rules from Table \ref{tab:weighted_averages}. We denote $(\xi_{n})_{1\le n \le N}$ the components of $x \in \mathbb{R}^N$. We set $\epsilon \in (0, +\infty)$ and, to circumvent the indefiniteness of the harmonic mean for vectors with a zero component, we define the \emph{leaky hyperbolic} function $h_{\epsilon}$ as
\begin{equation}
(\forall \xi \in \mathbb{R})\quad
h_{\epsilon}(\xi) = \begin{cases}
\displaystyle \frac{1}{\xi+\epsilon}-\epsilon & \mbox{if $\xi \in [0,1/\epsilon-\epsilon]$}\\
\displaystyle-\frac{\xi}{\epsilon^2} +\frac{1}{\epsilon}-\epsilon & \mbox{if $\xi < 0$}\\
\displaystyle-\epsilon^2 \Big(\xi-\frac{1}{\epsilon}+\epsilon\Big) & \mbox{if $\xi > 1/\epsilon-\epsilon$.}
\end{cases}
\end{equation}
Under these notations, Table \ref{tab:f_examples} summarizes the expressions for $f$, $f^{-1}$, as well as their associated definition domains, recovering the averaging rules from Table \ref{tab:weighted_averages}. Exact geometric and harmonic means formula are retrieved when $\epsilon$ goes to zero.

\begin{table*}[!ht]
    \centering
    \caption{Examples for $f$, $f^{-1}$, and definition domains, for $x = (\xi_{n})_{1\le n \le N}$. Geometric and harmonic formulas are retrieved when $\epsilon \to 0$.}
    \begin{adjustbox}{width=\textwidth}
    \begin{tabular}{l|c|c|c|c}
        Mean & $f(x)$ & $f$ domain & $f^{-1}(x)$ & $f^{-1}$ domain\\ \hline
        Arithmetic & $\operatorname{Id}$ & $[0,+\infty)^N$ & $\operatorname{Id}$ & $[0,+\infty)^N$ \Tstrut \\
        Geometric  & $\big(\ln(\xi_{n}+\epsilon) \big)_{1\le n \le N}$ & $[0,+\infty)^N$ & $\big(\exp(\xi_{n})-\epsilon\big)_{1\le n\le N}$ & $[\ln \epsilon,+\infty)^N$ \\
        Harmonic   & $\big(h_{\epsilon}(\xi_{n})\big)_{1\le n \le N}$ & $[0,+\infty)^N$ & $\big(h_{\epsilon}(\xi_{n})\big)_{1\le n \le N}$ & $(-\infty,\epsilon^{-1} - \epsilon]^N$ \\
        Power-$q$  & $(\xi_n^q)_{1\le n \le N}$ & $[0,+\infty)^N$ & $(\xi_n^{1/q})_{1\le n \le N}$ & $[0,+\infty)^N$\\
    \end{tabular}
     \end{adjustbox}
    \label{tab:f_examples}
\end{table*}

We now propose to extend the generalised average framework \eqref{e:genav} to the case when scalars $(\omega_{k})_{1\le k \le K}$ are replaced by matrices $(\Omega_{k})_{1\le k \le K}$ in $\mathbb{R}^{N\times N}$, so as to allow a full mixing of the weak learners. Given some functions $(f,f^{-1})$ defined as previously, we define the $f$-average output $\tilde{x} \in \mathbb{R}^N$ as
\begin{equation}\label{e:txfm1by}
\tilde{x} = f^{-1}\left(\sum_{k = 1}^K \Omega_k f(x_k)\right) = f^{-1}\big(W \boldsymbol{f}(\boldsymbol{x})\big).
\end{equation}
Hereabove, 
\begin{equation}
    W =[\Omega_1,\ldots,\Omega_K]
\in \mathbb{R}^{N \times KN},
\end{equation}
and function
\begin{align}
   \boldsymbol{f} & \colon \quad [0,+\infty)^{KN} \to C^{K} \qquad 
\\
& \quad \boldsymbol{x} = (x_k)_{1 \le k \le K}  \mapsto
\big(f(x_k)\big)_{1\le k \le K},
\end{align}
applies $f$ in a parallel (i.e., block-wise) manner to
the vector inputs
$(x_k)_{1 \le k \le K}$. 

In order to ensure the interpretability of the averaging operation in \eqref{e:txfm1by}, we propose to set
$W$ such that 
\begin{equation}
\begin{cases}
W \in [0, +\infty)^{N \times KN}, \\
W\mathds{1}_{KN} = \mathds{1}_{N}.
\end{cases}
\label{eq:consW}
\end{equation}
This guarantees, in particular, that $W \boldsymbol{f}(\boldsymbol{x})$ belongs to the definition domain of $f^{-1}$ (since this domain has been assumed to be convex). For instance, if $\boldsymbol{x} \in [0, 1]^{KN}$ (e.g., in a classification context), the constraint \eqref{eq:consW} on $W$ ensures that the output $\tilde{x}$ also belongs to $[0, 1]^N$. 

\begin{figure}[ht]
\centering
\begin{tikzpicture}[
	snode/.style={draw, rectangle, minimum size=0.5cm},
    scale=1.3, every node/.style={transform shape},
	]

\node        (input)                   {$\boldsymbol{x}$};
\node[snode] (f1)     [right=of input] {$\boldsymbol{f}$};
\node[snode] (W)      [right=of f1]    {$W$};
\node[snode] (f_inv)  [right=of W]     {$f^{-1}$};
\node        (output) [right=of f_inv] {$\tilde{x}$};

\draw[-latex] (input.east) -- (f1.west);
\draw[-latex] (f1.east) -- (W.west);
\draw[-latex] (W.east) -- (f_inv.west);
\draw[-latex] (f_inv.east) -- (output.west);
\end{tikzpicture}
\caption{\label{general_average_nn} Structure of the $f$-average network, i.e. a neural network that performs an $f$-average for ensembling}
\end{figure}
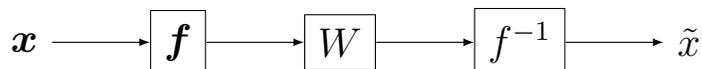

Remarkably, operation~\eqref{e:txfm1by}, that we call an $f$-average, can be represented as the application, on $\boldsymbol{x}$, of a two-layer neural network whose structure is drawn in Figure~\ref{general_average_nn}. This neural network is parametrized by the choice of $f$ (along with its inverse function $f^{-1}$) and by the weight matrix $W$. The former can be set by the user, while the latter can be determined through supervised learning, by minimizing a loss associated to the task at hand, over a training set. Once trained, the $f$-average network in Figure~\ref{general_average_nn} is interpretable, as the contribution of each output in the final prediction can easily be retrieved using the weights in matrix $W$, and the averaging operation is simply determined by the choice of $f$.

\subsection{Aggregated \texorpdfstring{$f$}{f}-averages}
\label{subsection:agg_f_average}

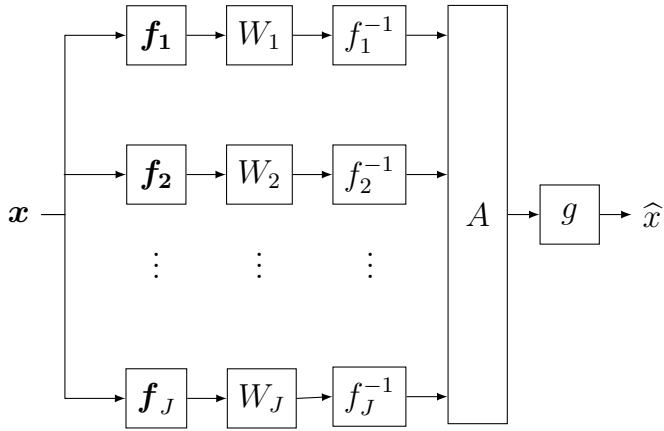
\begin{figure}[ht!]
    \centering
    \begin{tikzpicture}[
        snode/.style={draw, rectangle, minimum size=0.75cm},
        cnode/.style={draw, circle},
        rnode/.style={inner sep=0pt, text width=9mm},
        scale=1.05, every node/.style={transform shape},
        ]
    \node (x) {$\boldsymbol{x}$};
    \node[rnode] (r1)    [right=of x]      {};
    \node[snode] (f2)    [above=1mm of r1] {$\boldsymbol{f_2}$};
    \node[snode] (f1)    [above=of f2]     {$\boldsymbol{f_1}$};
    \node        (vdot1) [below=1mm of r1] {$\vdots$};
    \node[snode] (fK)    [below=of vdot1]  {$\boldsymbol{f}_J$};
    \node[rnode] (r2)    [right=0.5cm of r1]     {};
    \node[snode] (w1)    [right=0.5cm of f1]     {$W_1$};
    \node[snode] (w2)    [right=0.5cm of f2]     {$W_2$};
    \node        (vdot2) [right=0.9cm of vdot1] {$\vdots$};
    \node[snode] (wK)    [right=0.5cm of fK]     {$W_J$};
    \node[rnode] (r3)     [right=0.5cm of r2]     {};
    \node[snode] (f_inv1) [right=0.5cm of w1]     {$f^{-1}_1$};
    \node[snode] (f_inv2) [right=0.5cm of w2]     {$f^{-1}_2$};
    \node        (vdot3)  [right=1cm of vdot2] {$\vdots$};
    \node[snode] (f_invK) [below=2.05cm of f_inv2] {$f^{-1}_J$};
    \node[snode, minimum height=5.3cm] (a) [right=0.4cm of r3] {$A$};
    \node[snode] (g) [right=0.4cm of a] {$g$};
    \node (x_hat) [right=0.4cm of g] {$\widehat{x}$};
    \draw[-latex] (x.east) -- ++(0.3,0) |- (f1.west);
    \draw[-latex, shorten <=3mm] (f2-|x.east) -- (f2.west);
    \draw[-latex] (x.east) -- ++(0.3,0) |- (fK.west);
    \draw[-latex] (f1.east) -- (w1.west);
    \draw[-latex] (f2.east) -- (w2.west);
    \draw[-latex] (fK.east) -- (wK.west);
    \draw[-latex] (w1.east) -- (f_inv1.west);
    \draw[-latex] (w2.east) -- (f_inv2.west);
    \draw[-latex] (wK.east) -- (f_invK.west);
    \draw[-latex] (f_inv1.east) -- (f_inv1-|a.west);
    \draw[-latex] (f_inv2.east) -- (f_inv2-|a.west);
    \draw[-latex] (f_invK.east) -- (f_invK-|a.west);
    \draw[-latex] (a.east) -- (g.west);
    \draw[-latex] (g.east) -- (x_hat.west);
    \end{tikzpicture}
    \caption{\label{general_average_nn_j} Structure of the proposed aggregated $f$-average (AFA) neural network. It aggregates $J$ $f$-averages for ensembling, with $A \in [0, +\infty)^{N \times NJ}$. The activation function $g: \mathbb{R}^{N} \rightarrow \mathbb{R}^{N}$ is selected according to the task (e.g, softmax for classification, linear for regression).}
\end{figure}

The previous approach requires the prior choice for the average rule (i.e., $f$). To wave this restriction, we suggest aggregating $J > 1$ $f$-averages, associated to different functions $(f_j)_{1 \leq j \leq J}$. Resorting to the same structure as the one presented in the previous section, we define
\begin{equation}
(\forall j \in \{1,\ldots,J\}) \quad \tilde{x}_j = f_j^{-1} \Big( W_j \boldsymbol{f}_j(\boldsymbol{x}) \Big),
\label{e:ensembone0}
\end{equation}
where, for every $j\in \{1,\ldots,J\}$,
\begin{equation}
\begin{cases}
W_j \in [0, +\infty)^{N \times KN} \\
W_j\mathds{1}_{KN} = \mathds{1}_{N}$, $
\end{cases}
\label{eq:consWj}
\end{equation}
and $\boldsymbol{f}_j$ is a function operating component-wise from $[0,+\infty)^{KN}$ to $C_j$, associated to a given averaging function $f_j$, $f^{-1}_j$ is its inverse function operating component-wise from some convex set $C_j\subset \mathbb{R}^N$ to $[0,+\infty)^N$, and $\boldsymbol{x}$ concatenates columnwise the inputs $(x_k)_{1 \le k \le K}$. The resulting joint aggregate estimate of the $J$ outputs $(\tilde{x}_k)_{1 \leq j \leq J}$ is defined as
\begin{equation}
\widehat{x} = \sum_{j=1}^J A_{j} \tilde{x}_j = A \begin{bmatrix}
\tilde{x}_1\\
\vdots \\
\tilde{x}_J
\end{bmatrix},
\label{e:ensembone}
\end{equation}

\noindent with, for every $j\in \{1,\ldots,J\}$, $A_j \in [0, +\infty)^{N \times N}$, and $A \in [0, +\infty)^{N \times NJ}$ is the rowwise stacking of $(A_{j})_{1\le j \le J}$ matrices. Operations \eqref{e:ensembone0}-\eqref{e:ensembone} are equivalent to plug $\boldsymbol{x}$ as the input of a neural network with $J$ sub-networks (i.e., branches) of the form presented in Figure \ref{general_average_nn}, operating in parallel, followed by a linear layer involving the weight matrix $A$. We further propose to add a final activation function, $g\colon \mathbb{R}^{N} \to \mathbb{R}^{N}$, to control the domain of the output. For instance, a softmax activation can be used to get nonnegative outputs summing to one, in a classification context. The resulting network, called \emph{aggregated $f$-averages} (AFA), is displayed in Figure \ref{general_average_nn_j}. It has a limited number $JN^2(K + 1)$ of linear parameters, namely the entries of matrices $W_1, \ldots, W_J$, and $A$. The training of these parameters can follow a classical supervised learning approach. Given a sample and its ground truth, the task model loss is computed (e.g., a cross-entropy loss for classification), before updating the weights from all layers using a backpropagation algorithm (e.g., Adam optimizer \cite{kingma2014adam}). Constraints \eqref{eq:consWj} on weight matrices $(W_j)_{1 \leq j \leq J}$ can simply be imposed through a projection step of each row of these matrices on the convex unit simplex set \cite{CondatSimplex}, performed after each backpropagation update. Let us remark that this model maintains the interpretability properties of the $J$ $f$-average sub-models it contains. Furthermore, weights from matrix $A$ can be viewed as the contribution level of each type of average model.

\subsection{Application of Aggregated \texorpdfstring{$f$}{f}-averages to FSCIL}
\label{sec:few_shot_ensembling}

We now present our approach, to address FSCIL, using ensemble learning by AFA. We propose to consider FSCIL as a combination of successive few-shot learning problems. To do so, for each session, we train a few-shot learning classifier. Each of these weak classifiers can be viewed as a model specialised on its own session data. This enables to build a diverse set of models which are then gathered by our AFA ensembling model. By combining predictions from every weak classifiers, the ensembling model will be able to output a prediction for all classes from base to current sessions. Let us now describe the detailed architecture, and our specific strategy for padding unseen classes predictions.  

\subsubsection{Architecture of AFA for FSCIL}
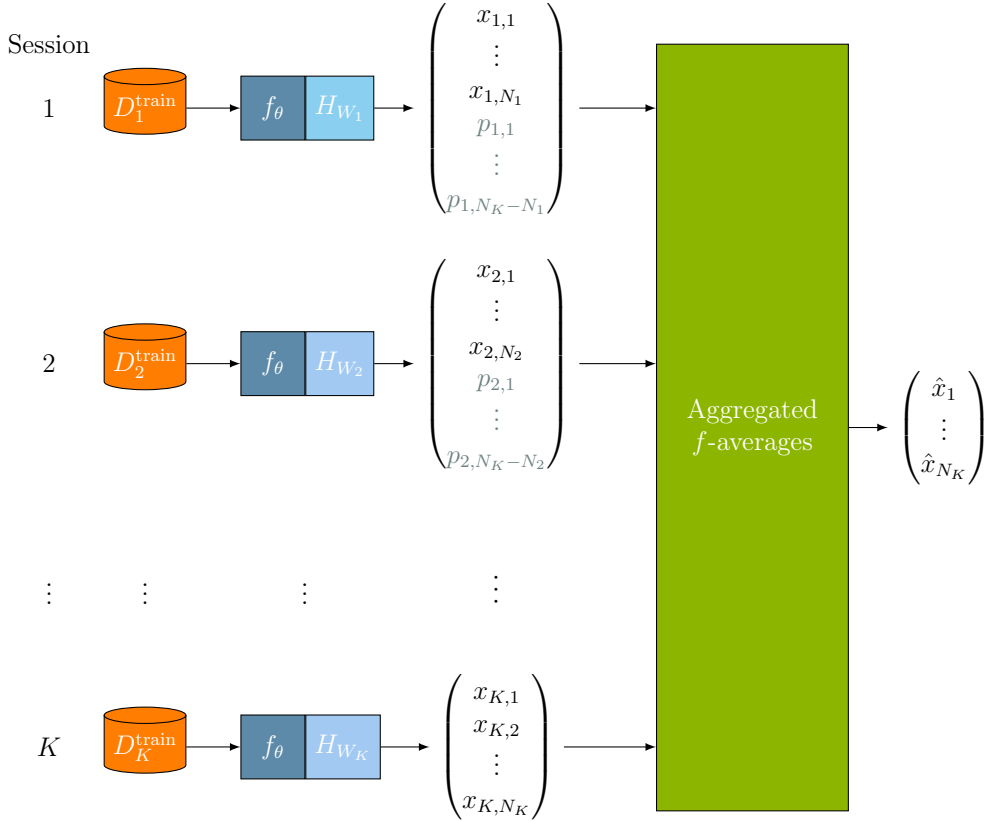
\begin{figure}[ht!]
    \centering
    \begin{tikzpicture}[
            snode/.style={rectangle, draw, minimum width=1cm, minimum height=1cm, text=white},
            dnode/.style={cylinder, draw, aspect = 0.2, shape border rotate = 90, text=white},
            scale=0.85, every node/.style={transform shape}
        ]

        \node (s0) at (-11,6)  {Session};
        \node (s1) at (-11,5)  {1};
        \node (s2) at (-11,1)  {2};
        \node (s3) at (-11,-2.5) {$\vdots$};
        \node (s4) at (-11,-5) {$K$};
        \node[dnode, fill=amber] (d1) at (-9.5,5)  {$D_1^\text{train}$};
        \node[dnode, fill=amber] (d2) at (-9.5,1)  {$D_2^\text{train}$};
        \node[]      (d3) at (-9.5,-2.5) {$\vdots$};
        \node[dnode, fill=amber] (d4) at (-9.5,-5) {$D_K^\text{train}$};
        \node[snode, fill=airforceblue] (m1) at (-7.5,5)  {$f_\theta$};
        \node[snode, fill=babyblue] (m1h) [right=0cm of m1, align=center]  {$H_{W_1}$};
        \node[snode, fill=airforceblue] (m2) at (-7.5,1)  {$f_\theta$};
        \node[snode, fill=babyblueeyes] (m2h) [right=0cm of m2, align=center]  {$H_{W_2}$};
        \node[]      (m3) at (-7,-2.5) {$\vdots$};
        \node[snode, fill=airforceblue] (m4) at (-7.5,-5) {$f_\theta$};
        \node[snode, fill=babyblueeyes] (m4h) [right=0cm of m4, align=center]  {$H_{W_K}$};

        \node[] (o1) at (-4,5) {
            $\begin{pmatrix}
                x_{1,1} \\ 
                \vdots \\ 
                x_{1, N_1} \\
                \textcolor{aurometalsaurus}{p_{1,1}} \\
                \textcolor{aurometalsaurus}{\vdots} \\
                \textcolor{aurometalsaurus}{p_{1, N_K - N_1}} \\
            \end{pmatrix}$
        };
        \node[] (o2) at (-4,1) {
            $\begin{pmatrix}
                x_{2,1} \\ 
                \vdots \\ 
                x_{2, N_2} \\
                \textcolor{aurometalsaurus}{p_{2,1}} \\
                \textcolor{aurometalsaurus}{\vdots} \\
                \textcolor{aurometalsaurus}{p_{2, N_K - N_2}} \\
            \end{pmatrix}$
        };
        \node[] (o3) at (-4,-2.5) {\rotatebox{90}{$\cdots$}};
        \node[] (o4) at (-4,-5) {
            $\begin{pmatrix}
                x_{K,1} \\ 
                x_{K,2} \\ 
                \vdots \\ 
                x_{K, N_K} \\
            \end{pmatrix}$
        };
        \node[snode, minimum height=12cm, minimum width=3cm, align=center, fill=applegreen] (e) at (0,0) {Aggregated \\ $f$-averages};
        \node (x) at (3,0) {
            $\begin{pmatrix}
                \hat{x}_{1} \\
                \vdots            \\
                \hat{x}_{N_K}
            \end{pmatrix}$
        };

        \draw[-latex] (d1.east) -- (m1.west);
        \draw[-latex] (d2.east) -- (m2.west);
        \draw[-latex] (d4.east) -- (m4.west);
        \draw[-latex] (m1h.east) -- (o1.west);
        \draw[-latex] (m2h.east) -- (o2.west);
        \draw[-latex] (m4h.east) -- (o4.west);
        \draw[-latex] (o1.east) -- (o1-|e.west);
        \draw[-latex] (o2.east) -- (o2-|e.west);
        \draw[-latex] (o4.east) -- (o4-|e.west);
        \draw[-latex] (e.east) -- (x.west);
    \end{tikzpicture}
    \caption{AFA ensembling for FSCIL. The CNN feature extractor $f_\theta$ is trained in base session and frozen for following sessions. Its classification head $H_{W_1}$ is a standard classification layer. For subsequent sessions (for $k \in \{2,\ldots,K\}$) SimpleShot classifications heads $H_{W_k}$ are used, based on nearest-neighbour classifiers, and fine-tuned using $D_k^\text{train}$. A padding $p$ is added to compensate the size of outputs $x$ from classification heads.}
    \label{fig:FSCIL}
\end{figure}

The complete architecture is displayed in Fig.~\ref{fig:FSCIL}. First, and similarly to \cite{chen2019closer, tao2020fscil}, we consider a CNN classification model (here, a ResNet-18), as the composition of a feature extractor $f_\theta$ parametrised by some network weights gathered in a vector $\theta$, and a classification head $H_{W_k}$ with $W_k$ its weights for session $k$. The CNN is trained during base session ($k=1$), namely the only session with a substantial amount of data. The CNN's backbone $f_\theta$ (dark blue box in Fig.~\ref{fig:FSCIL}) is frozen, and will serve the purpose of a feature extractor for following sessions ($k \in \{2,\ldots,K \}$). Its last layer, i.e., the classification head $H_{W_1}$ (top light blue box in Fig.~\ref{fig:FSCIL}) is a weak classifier specialised on the base session.

\newpage
Hence, for $k \in \{2,\ldots,K \}$, we replace it by new classification heads $H_{W_k}$. Specifically, instead of a CNN last layer, we rely on the few-shot learning method SimpleShot \cite{simpleshot}. SimpleShot is a nearest-neighbour classifier which, given a session $k$, determines an estimated class $\widehat{c}(i)$ of a test image $i \in D_k^\text{test}$ by retrieving the closest mean centroid in the feature space
\begin{equation}
    \widehat{c}(i) = \argmin_{c \in C_k^\text{test}} \|z - \bar{z_c}\|^2,
\end{equation}
where $z$ is the $\ell_2$-normalised feature vector of a sample image $i$ 
\begin{equation}
    z = \frac{f_\theta(i)}{\|f_\theta(i)\|_2}
\end{equation}
and $\{\bar{z_c} \mid c \in C_k^\text{test}$\} are the mean centroid of each class
\begin{equation}
 (\forall k \in \{1, \dots, K\}) 
 (\forall c \in C_k^\text{train}) 
    \quad \bar{z_c} = \frac{1}{n_{\text{shots}}}\sum_{n = 1}^{n_{\text{shots}}} z_{n,c}.
\end{equation}

Hence, the SimpleShot classification head coupled with the feature extractor forms the weak classifier for each session $k \in \{2,\ldots,K\}$. All these weak classifiers should perform well on images whose classes belong to their respective session since they are trained only on these classes. However, they might not generalise on (unseen) classes from other sessions. To overcome this issue, we propose to ensemble all weak classifiers outputs, using our AFA model (orange box in Fig.~\ref{fig:FSCIL}), so as to produce a final model capable of predicting any class. At session $k$, input of the ensembling model consists of $k$ outputs, each from a different weak classifier. However, except the CNN from base session, weak classifiers output distances and not probabilities. To convert them, we use the following formula:

\begin{equation}\label{eq:simpleshot_probability}
    x = \textsf{softmin}(\tau \|z - \bar{z_c}\|^2)
\end{equation}

\noindent where $\tau>0$ is a learnable temperature scaling parameter. Since SimpleShot relies on the smallest distance to predict classes, we need to transform low distances into high probabilities, hence the \textsf{softmin} function. Once learnt, the temperature parameter scales the output distance so that the \textsf{softmin} operation produces meaningful numerical values. 

\subsubsection{Padding and inlierness}
As it can be seen in Fig.~\ref{fig:FSCIL}, once session $K$ is reached, we obtain $K$ weak classifiers outputs, with a vector of prediction of a different size $N_k ~ (\forall k \in \{1, \dots ,K\})$ (see also Eq. (\ref{eq:classFSCIL})). However, our AFA ensembling only averages inputs of the same length. To cope with this situation, we must add a padding at the end of each prediction vector (except, obviously, for the last one which is already of size $N_K$) in order to obtain $K$ prediction vectors which are all of the same size, i.e. predicting the same number of classes, namely the number of classes $N_K$ of session $K$ (cf. Eq.~(\ref{eq:padding})). The size of the padding is $N_K - N_k$, and $(p_i)_{1 \leq i \leq N_K-N_k}$ denotes the padding vector so that
\begin{equation}\label{eq:padding}
    x = \begin{pmatrix}
         x_1 \\ 
         x_2 \\ 
         \vdots \\ 
         x_{N_k} \\
         p_1 \\
         p_2 \\
         \vdots \\
         p_{N_K - N_k} \\
    \end{pmatrix}
\end{equation}
We initially set the components of $p$ to be all equal to $p_i = 1 / (N_K - N_k)$, considering an uniform probability among the remaining classes. 
However, by doing so, $x$ in Eq.~\eqref{eq:padding} becomes the concatenation of two probabilities vector, not summing to one, with an obvious lack of interpretability. We thus propose a corrective approach, where we multiply the predictions $(x_i)_{1 \leq i \leq N_k}$ and the padded values $(p_i)_{1 \leq i \leq N_K - N_k}$ by appropriate scaling factors, ensuring both normalization and meaningfulness of the output. To do so, we design an inlierness indicator assessing whether a given input is regarded as probably being in the set of learnt classes of that particular weak classsifier or not. The inlierness indicator relies on the quantity $\max\limits_{1 \le i \le N_k}x_i$, which indeed indicates the weak classifier level of confidence of the prediction $x$ from Eq.~(\ref{eq:simpleshot_probability}).
If the latter value is higher than a certain threshold $t>0$ (set as an hyperparameter), then the prediction is considered to be reliable and likely to belong to the set of learnt classes for this particular weak classifier. In such a case, the normalisation factor reflects the confidence by increasing the weight of corresponding predictions $x_i$ with respect to padding values $p_i$. Otherwise, if the prediction confidence is lower than the threshold, the normalisation factor emphasizes padding values instead of the predictions. More precisely, new prediction values $x'_i$ are computed as follows, for every $i \in \{1,\ldots,N_k\}$,
\begin{equation}
    x'_i = \begin{cases}
      \frac{1}{2} (\frac{N_k}{N_K} + 1)  x_i, & \text{if} \max\limits_{1 \le i \le N_k}x_i \geq t\\
      \frac{1}{2} (\frac{N_k}{N_K}) x_i, & \text{otherwise}.
    \end{cases}
\end{equation}

\noindent Furthermore, the corrected values $(p'_i)_{1 \leq i \leq N_K - N_k}$ are computed using the complimentary probability, in order to obtain a prediction vector summing to one. For every $i \in \{1,\ldots,N_K-N_k\}$,
\begin{equation}
    p'_i = \begin{cases}
      (1 - \frac{1}{2} (\frac{N_k}{N_K} + 1)) p_i, & \text{if} \max\limits_{1 \le i \le N_k} x_i \ge t\\
      (1 - \frac{1}{2} (\frac{N_k}{N_K})) p_i, & \text{otherwise},
    \end{cases}
\end{equation}
where we recall that $p_i = 1/(N_K-N_k)$ for all $i \in \{1,\ldots,N_K-N_k\}$.

Finally, to train the AFA ensembling model, we adopt a \emph{prototype rehearsal} method similarly to \cite{rebuffi2017icarl}, i.e., weak classifiers perform predictions on all training prototypes used during all sessions in order to provide training inputs for our final ensembling model.

\section{Experiments}
We now present our experimental results, with the aim to show (i) the validity of the combined FSCIL and ensembling pipeline, (ii) the benefits of our AFA method when compared to state-of-the-art ensemble learning methods, in that context.  

We conduct experiments on four image classification datasets typically used in few-shot or incremental learning, namely mini-ImageNet \cite{matching_net}, CUB-200 \cite{cub200}, tiered-ImageNet \cite{tiered_imagenet}, and FGVC-Aircraft \cite{fgvc-aircraft}. Let us first describe each dataset specificity.

\paragraph{mini-ImageNet} Based on ImageNet \cite{imagenet}, this dataset provides 500 training samples and 100 test samples for each of the 100 classes. We replicate \cite{tao2020fscil} setting parameters by having 9 sessions with a number of base classes $n_\text{class\_base}$ of 60 and 5-\emph{way} 5-\emph{shots} for following sessions.

\paragraph{tiered-ImageNet} Also based on ImageNet, this dataset provides 506 classes of images. This allows us to evaluate our method on a larger scale than what mini-ImageNet allows us to. To do so, we set the total number of sessions to 11 with a number of base classes $n_\text{class\_base}$ of 100 and 10-\emph{way} 5-\emph{shots} for following sessions, similarly to the CUB-200 dataset.

\paragraph{CUB-200} As opposed to ImageNet (and its smaller versions), this dataset is devised for fine-grained image classification. For this dataset too, we reproduce \cite{tao2020fscil} setting with 11 sessions, a number of base classes $n_\text{class\_base}$ of 100 and 10-\emph{way} 5-\emph{shots} for following sessions.

\paragraph{FGVC-Aircraft} To complete previous datasets, we add FGVC-Aircraft as it is another fine-grained image classification dataset, with a similar size compared to mini-ImageNet. We set it up similarly to the latter with 10 sessions, a number of base classes $n_\text{class\_base}$ of 50 and 5-\emph{way} 5-\emph{shots} for following sessions.

These four datasets allow us to provide experimental results for standard and fine-grained image classification and, for both, with a smaller and a larger setting size. 

For every experiment, we use ResNet-18 \cite{he2016deep} as base model. Trained on base session, we set the mini-batch size to 128, the learning rate to 0.1 and the momentum to 0.9. The learning rate is decreased by a factor 10 on epoch 30 and again on epoch 40.
For subsequent sessions, its classification head is removed and replaced as described in Section \ref{sec:few_shot_ensembling} and feature extractor weights are frozen. Our method is implemented using Pytorch \cite{NEURIPS2019_9015} and models were trained on a Nvidia Tesla V100 GPU. 

Experimental results of our proposed AFA pipeline, are compared against state-of-the-art methods dedicated to FSCIL, namely SPPR \cite{zhu2021self}, CEC \cite{zhang2021few}, and FACT \cite{zhou2022forward}. We also display results from the incremental learning methods LUCIR \cite{hou2019learning} and EEIL \cite{castro2018end}, for reference. Regarding ensemble learning benchmarks, we perform comparisons with standard output fusion strategies, namely classic averaging rules (arithmetic, geometric, and harmonic) and a majority vote scheme. In addition, stacked NNs performing the output fusion are also experimented (a shallow NN, a deep NN, and a weighted average NN). 

We first discuss in Section~\ref{sec:metric} the choice for the comparative metrics, which is particularly challenging in the FSCIL context. We then provide our experimental results and discuss those in Section~\ref{sec:expe}.

\subsection{Setting comparison metrics}
\label{sec:metric}

\begin{table*}[!ht]
    \centering
    \begin{adjustbox}{width=\textwidth}
        \begin{tabular}{l|cccc|cccc}
            ~ & \multicolumn{4}{c}{mini-ImageNet} & \multicolumn{4}{|c}{CUB-200} \\
            Method & mean acc. & acc. base & acc. new & F1 & mean acc. & acc. base & acc. new & F1 \\ \hline
            No fine-tuning & 39.26 & 65.43 & 0.00 & 30.08 & 33.33 & 67.42 & 0.00 & 24.64 \\
            FACT \cite{zhou2022forward} & 49.51 & 73.80 & 13.07 & 44.77 & 56.43 & 72.93 & 40.30 & 56.03 \\
            AFA & 48.11 & 66.96 & 27.80 & 48.75 & 52.22 & 70.53 & 34.33 & 51.58 \\
        \end{tabular}     
    \end{adjustbox}
    \caption{Detailed metrics on last session of mini-ImageNet and CUB-200, for 3 compared methods. Here, the model `No fine-tuning', only trained on base session and ignoring new sessions displays a very close mean accuracy over all classes, hiding the fact that it can hardly handle new classes. This table illustrates that F1 score better reflects performance in an imbalanced classification setting.}
    \label{tab:metrics_comparison}
\end{table*}

Let us first discuss the comparison metrics, used in our subsequent experiments. In most experimental settings, $n_\text{way}$ being much smaller than $n_\text{class\_base}$, the FSCIL problem is an imbalanced classification one. While the imbalance is especially strong in earlier sessions due to the very few novel classes compared to the bigger base training set, this can also be observed when reaching last session where $n_\text{class}$ can be inferior to $n_\text{class\_base}$. Preliminary experiments show that some methods might reach a fairly strong mean accuracy over all (base \emph{and} new) classes by focusing only on base classes. While this can prevent efficiently catastrophic forgetting, it can also be counterproductive in the learning of new classes. Table~\ref{tab:metrics_comparison} illustrates this behaviour with an extreme example. 

For example, we devise a model called `No fine-tuning' trained on base session only, i.e. that ignored following sessions training. For those sessions, vector predictions are completed with zeros as the model does not recognize any class beyond base session. We can observe that such model retains a decent mean accuracy thanks to its performance on base session classes. In contrast, both FACT and our AFA appear to generalize well from the base to the new session. The bias of the mean accuracy metric is well-known \cite{branco2016survey}. Instead, in such imbalanced settings, the F1 score is preferred, defined as
\begin{equation}
\text{F1} = \frac{2 \times \text{Precision} \times \text{Recall}}{\text{Precision + Recall}}  
\end{equation}

\noindent where Precision = TruePositive / (TruePositive + FalsePositive) and Recall = TruePositive / (TruePositive + FalseNegative). As Table \ref{tab:metrics_comparison} shows, imbalanced performance between base and new classes are better accounted for with this metric than with average accuracy over all classes. Following results will be reported using only F1 score, based on this discussion. In all the forthcoming tables, best (i.e., highest) F1 scores among all benchmarks are highlighted in bold, while the top F1 scores among ensembling methods are underlined. 

\subsection{Experimental results}
\label{sec:expe}

\begin{table}[!ht]
    \centering
    \begin{adjustbox}{width=\textwidth}
    \begin{tabular}{llccccccccccc}
        &  & \multicolumn{9}{c}{mini-ImageNet} & \\
        \multirow{2}*{Strategy} & \multirow{2}*{Method} & \multicolumn{9}{c}{F1 score (\%) for each session} \\
        \cmidrule(lr){3-11}
        & & 0 & 1 & 2 & 3 & 4 & 5 & 6 & 7 & 8 \\
        \cmidrule(lr){1-11}
        \multirow{2}*{Incremental} & LUCIR \cite{hou2019learning} & \textbf{75.93} & 74.21 & 56.88 & 44.02 & 31.86 & 25.83 & 22.47 & 18.48 & 14.75 \\ 
        & EEIL \cite{castro2018end} & 70.92 & 69.93 & 5.64 & 1.63 & 1.88 & 2.60 & 2.74 & 1.27 & 1.05 \\ 
        \cmidrule(lr){1-11}
        & SPPR \cite{zhu2021self} & 62.40 & 62.60 & 57.64 & 52.90 & 49.01 & 45.75 & 42.81 & 39.34 & 36.90 \\ 
        FSCIL & CEC \cite{zhang2021few} & 72.12 & 66.95 & 62.94 & 59.63 & 56.91 & 53.87 & 51.37 & 49.36 & 47.72 \\ 
        & FACT \cite{zhou2022forward} & 75.37 & 70.49 & 65.99 & 62.21 & \textbf{58.89} & 55.71 & 52.50 & \textbf{50.40} & 48.59 \\
        \cmidrule(lr){1-11}
        & Arithmetic mean & 75.37 & 69.51 & 64.58 & 60.05 & 55.43 & 51.75 & 48.56 & 45.26 & 42.74 \\ 
        & Geometric mean & 75.37 & 69.98 & 65.27 & 59.87 & 55.41 & 51.13 & 47.93 & 44.23 & 41.26 \\ 
        & Harmonic mean & 75.37 & \underline{\textbf{70.82}} & \underline{\textbf{66.72}} & 62.21 & 57.53 & 52.44 & 48.64 & 44.66 & 41.26 \\ 
        Ensemble & Majority vote & 75.37 & 67.52 & 62.50 & 58.29 & 53.36 & 50.08 & 46.69 & 43.94 & 40.88 \\ 
        Learning & Shallow NN & 75.37 & 65.93 & 63.90 & 59.57 & 56.77 & 54.35 & 50.73 & 43.81 & 46.67 \\ 
        & Deep NN & 75.37 & 60.55 & 61.12 & 59.26 & 55.68 & 53.51 & 46.96 & 47.04 & 43.77 \\ 
        & Weighted avg. & 75.37 & 64.24 & 62.82 & 53.77 & 53.17 & 50.58 & 47.50 & 44.18 & 37.46 \\ 
        & AFA & 75.37 & 68.83 & 66.54 & \underline{\textbf{63.09}} & \underline{57.86} & \underline{\textbf{56.40}} & \underline{\textbf{53.50}} & \underline{50.36} & \underline{\textbf{48.80}}\\ 
    \end{tabular}
    \end{adjustbox}
    \caption{Benchmark on mini-ImageNet. Methods from various strategies are included: incremental learning, FSCIL and ensemble learning.}
    \label{tab:f1_sota_min}
\end{table}

\begin{table}[!ht]
    \centering
    \begin{adjustbox}{width=\textwidth}
    \begin{tabular}{llccccccccccc}
        &  & \multicolumn{9}{c}{CUB-200} & \\
        \multirow{2}*{Strategy} & \multirow{2}*{Method} & \multicolumn{11}{c}{F1 score (\%) for each session} \\
        \cmidrule(lr){3-13}
        & & 0 & 1 & 2 & 3 & 4 & 5 & 6 & 7 & 8 & 9 & 10\\
        \cmidrule(lr){1-13}
        ~ & LUCIR \cite{hou2019learning} & 72.13 & 72.90 & 65.74 & 60.38 & 57.06 & 49.51 & 44.17 & 41.29 & 40.01 & 37.57 & 37.27 \\ 
        ~ & EEIL \cite{castro2018end} & 68.40 & 68.44 & 51.27 & 43.59 & 39.45 & 36.22 & 34.12 & 31.27 & 29.19 & 27.10 & 26.47 \\
        \cmidrule(lr){1-13}
        ~ & SPPR \cite{zhu2021self} & 68.66 & 63.18 & 58.70 & 54.44 & 51.67 & 47.92 & 46.22 & 45.59 & 42.34 & 40.35 & 38.74 \\ 
        FSCIL & CEC \cite{zhang2021few} & 75.69 & 71.56 & 67.94 & 63.15 & 62.22 & 57.95 & 57.26 & 55.61 & 54.49 & 53.30 & 52.12 \\ 
        ~ & FACT \cite{zhou2022forward} & \textbf{76.75} & \textbf{73.53} & \textbf{70.34} & \textbf{65.39} & \textbf{64.79} & \textbf{61.54} & \textbf{61.07} & \textbf{59.43} & \textbf{57.49} & \textbf{57.30} & \textbf{56.29} \\
        \cmidrule(lr){1-13}
        ~ & Arithmetic mean & 75.70 & 62.01 & \underline{63.93} & \underline{61.47} & 58.19 & 55.74 & 53.73 & 52.63 & 50.40 & 48.59 & 47.22 \\ 
        ~ & Geometric mean & 75.70 & 52.08 & 51.77 & 50.58 & 49.82 & 48.56 & 48.37 & 47.44 & 46.79 & 45.62 & 44.82 \\ 
        ~ & Harmonic mean & 75.70 & 52.17 & 50.64 & 47.17 & 45.26 & 43.57 & 42.24 & 41.10 & 39.33 & 38.59 & 37.27 \\ 
        Ensemble & Majority vote & 75.70 & 58.64 & 61.42 & 60.39 & 57.92 & 54.05 & 51.42 & 49.03 & 47.56 & 45.38 & 44.29 \\ 
        Learning & Shallow NN & 75.70 & 55.29 & 57.19 & 60.17 & 59.84 & 57.98 & \underline{57.28} & \underline{55.08} & \underline{53.85} & \underline{53.62} & 50.57 \\ 
        ~ & Deep NN & 75.70 & 49.09 & 57.48 & 57.72 & 55.00 & 53.15 & 47.52 & 46.94 & 42.41 & 41.83 & 38.43 \\ 
        ~ & Weighted avg. & 75.70 & 36.73 & 40.86 & 56.16 & 57.00 & 52.25 & 49.97 & 43.64 & 49.60 & 39.21 & 46.52 \\ 
        ~ & AFA & 75.70 & \underline{63.08} & 61.53 & 61.13 & \underline{60.01} & \underline{57.99} & 56.13 & 53.79 & 51.95 & 51.45 & \underline{51.58} \\ 
    \end{tabular}
    \end{adjustbox}
    \caption{Benchmark on CUB-200. Methods from various strategies are included: incremental learning, FSCIL and ensemble learning.}
    \label{tab:f1_sota_cub}
\end{table}

We now provide all our comparative results, namely the performance metrics for standard datasets mini-ImageNet and CUB-200 in Tables \ref{tab:f1_sota_min} and \ref{tab:f1_sota_cub}, and performance curves for tiered-ImageNet and FGVC-Aircraft in Figure~\ref{fig:f1_sota_tiered_aircraft}. As already mentioned, this variety of datasets allows us to benchmark the methods on standard image classification datasets (mini-ImageNet and tiered-ImageNet) and on fine-grained image classification datasets (FGVC-Aircraft and CUB-200). It also allows us to compare differences between smaller (100 classes in total for mini-ImageNet and FGVC-Aircraft) and larger (200 classes in total for tiered-ImageNet and CUB-200) datasets.

\begin{figure*}
	\centering
        \begin{adjustbox}{width=\textwidth}
		\begin{tikzpicture}
        \begin{groupplot}[
                    group style={group size= 2 by 1},
                    xtick=data,
                    xlabel=Session,
        ]
        \nextgroupplot[title=FGVC-Aircraft, ylabel={F1 score (\%)}]
        \addplot[yellow, mark=pentagon] table[x=Session, y=EEIL]{aircraft_f1.dat};
        \label{plot:EEIL}
        \addplot[purple, mark=o] table[x=Session, y=LUCIR]{aircraft_f1.dat};
        \label{plot:LUCIR}
        \addplot[orange, mark=diamond] table[x=Session, y=SPPR]{aircraft_f1.dat};
        \label{plot:SPPR}
        \addplot[red, mark=triangle] table[x=Session, y={CEC}]{aircraft_f1.dat};
        \label{plot:CEC}
        \addplot[green, mark=square] table[x=Session, y={FACT}]{aircraft_f1.dat};
        \label{plot:FACT}
        \addplot[blue, mark=*] table[x=Session, y={Agg. $f$-avg}]{aircraft_f1.dat};
        \label{plot:ours}
        \coordinate (top) at (rel axis cs:0,1);
        
        \nextgroupplot[title=tiered-ImageNet]
        \addplot[yellow, mark=pentagon] table[x=Session, y=EEIL]{tiered_imagenet_f1.dat};
        \addplot[purple, mark=o] table[x=Session, y=LUCIR]{tiered_imagenet_f1.dat};
        \addplot[orange, mark=diamond] table[x=Session, y=SPPR]{tiered_imagenet_f1.dat};
        \addplot[red, mark=triangle] table[x=Session, y={CEC}]{tiered_imagenet_f1.dat};
        \addplot[green, mark=square] table[x=Session, y={FACT}]{tiered_imagenet_f1.dat};
        \addplot[blue, mark=*] table[x=Session, y={Agg. $f$-avg}]{tiered_imagenet_f1.dat};
        \coordinate (bot) at (rel axis cs:1,0);
    \end{groupplot}

\path (top|-current bounding box.north)--
      coordinate(legendpos)
      (bot|-current bounding box.north);
\matrix[
    matrix of nodes,
    anchor=south,
    draw,
    inner sep=0.2em,
    draw
  ]at([yshift=-51ex]legendpos)
  {
    \ref{plot:EEIL}& EEIL \cite{castro2018end} &[5pt]
    \ref{plot:LUCIR}& LUCIR \cite{hou2019learning} &[5pt]
    \ref{plot:SPPR}& SPPR \cite{zhu2021self} &[5pt]
    \ref{plot:CEC}& CEC \cite{zhang2021few} &[5pt]
    \ref{plot:FACT}& FACT \cite{zhou2022forward} &[5pt]
    \ref{plot:ours}& AFA \\};
\end{tikzpicture}
\end{adjustbox}
	\caption{F1 comparison with the state of the art on FGVC-Aircraft and tiered-ImageNet}
 \label{fig:f1_sota_tiered_aircraft}
\end{figure*}
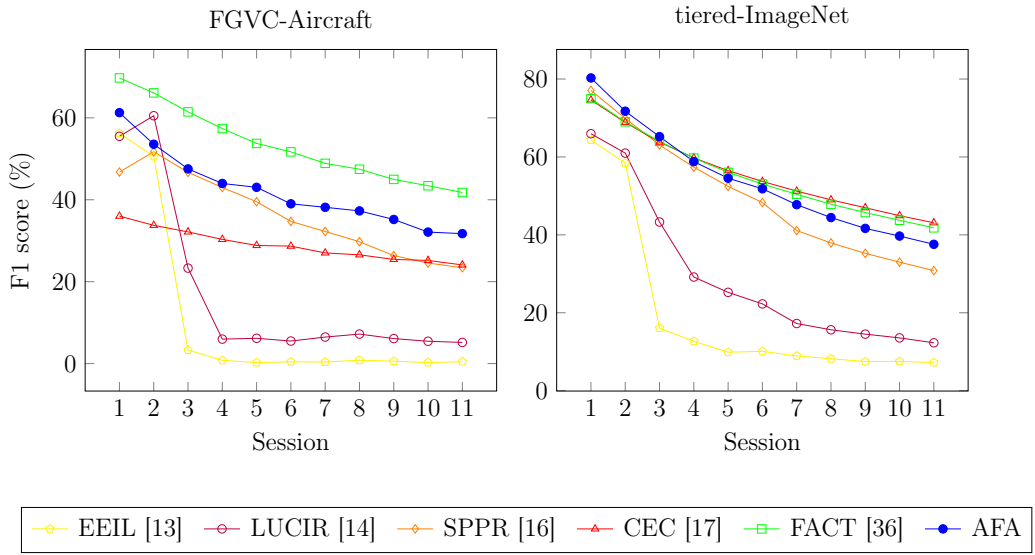

\begin{figure*}
	\centering
        \begin{adjustbox}{width=\textwidth}
		\begin{tikzpicture}
            \begin{groupplot}[
                group style={group size= 2 by 2, vertical sep=1.3cm},
                xtick=data,
                ymajorgrids=true,
                grid style=dashed,
            ]
        \nextgroupplot[title=mini-ImageNet, ylabel={F1 score (\%)}]
        \addplot[yellow, mark=pentagon] table[x=Session, y={Shallow NN}]{mini_imagenet_f1.dat};
        \addplot[pink, mark=pentagon] table[x=Session, y={Deep NN}]{mini_imagenet_f1.dat};
        \addplot[cyan, mark=pentagon] table[x=Session, y={Weighted average NN}]{mini_imagenet_f1.dat};
        \addplot[purple, mark=o] table[x=Session, y=arithmetic]{mini_imagenet_f1.dat};
        \addplot[orange, mark=diamond] table[x=Session, y=geometric]{mini_imagenet_f1.dat};
        \addplot[red, mark=triangle] table[x=Session, y={harmonic}]{mini_imagenet_f1.dat};
        \addplot[green, mark=square] table[x=Session, y={majority}]{mini_imagenet_f1.dat};
        \addplot[blue!80, mark=*] table[x=Session, y=Ours]{mini_imagenet_f1.dat};
        \coordinate (top) at (rel axis cs:0,1);
        
        \nextgroupplot[title=FGVC-Aircraft]
        \addplot[yellow, mark=pentagon] table[x=Session, y={Shallow NN}]{aircraft_f1.dat};
        \label{plot:shallow}
        \addplot[pink, mark=pentagon] table[x=Session, y={Deep NN}]{aircraft_f1.dat};
        \label{plot:deep}
        \addplot[cyan, mark=pentagon] table[x=Session, y={Weighted average NN}]{aircraft_f1.dat};
        \label{plot:wa}
        \addplot[purple, mark=o] table[x=Session, y=arithmetic]{aircraft_f1.dat};
        \label{plot:arithmetic}
        \addplot[orange, mark=diamond] table[x=Session, y=geometric]{aircraft_f1.dat};
        \label{plot:geometric}
        \addplot[red, mark=triangle] table[x=Session, y={harmonic}]{aircraft_f1.dat};
        \label{plot:harmonic}
        \addplot[green, mark=square] table[x=Session, y={majority}]{aircraft_f1.dat};
        \label{plot:majority}
        \addplot[blue!80, mark=*] table[x=Session, y={Agg. $f$-avg}]{aircraft_f1.dat};
        \label{plot:ours2}

        \nextgroupplot[title=CUB-200, xlabel=$K$, ylabel={F1 score (\%)}]
        \addplot[yellow, mark=pentagon] table[x=Session, y={Shallow NN}]{cub_f1.dat};
        \addplot[pink, mark=pentagon] table[x=Session, y={Deep NN}]{cub_f1.dat};
        \addplot[cyan, mark=pentagon] table[x=Session, y={Weighted average NN}]{cub_f1.dat};
        \addplot[purple, mark=o] table[x=Session, y=arithmetic]{cub_f1.dat};
        \addplot[orange, mark=diamond] table[x=Session, y=geometric]{cub_f1.dat};
        \addplot[red, mark=triangle] table[x=Session, y={harmonic}]{cub_f1.dat};
        \addplot[green, mark=square] table[x=Session, y={majority}]{cub_f1.dat};
        \addplot[blue!80, mark=*] table[x=Session, y=Ours]{cub_f1.dat};
        
        \nextgroupplot[title=tiered-ImageNet, xlabel=$K$]
        \addplot[yellow, mark=pentagon] table[x=Session, y={Shallow NN}]{tiered_imagenet_f1.dat};
        \addplot[pink, mark=pentagon] table[x=Session, y={Deep NN}]{tiered_imagenet_f1.dat};
        \addplot[cyan, mark=pentagon] table[x=Session, y={Weighted average NN}]{tiered_imagenet_f1.dat};
        \addplot[purple, mark=o] table[x=Session, y=arithmetic]{tiered_imagenet_f1.dat};
        \addplot[orange, mark=diamond] table[x=Session, y=geometric]{tiered_imagenet_f1.dat};
        \addplot[red, mark=triangle] table[x=Session, y={harmonic}]{tiered_imagenet_f1.dat};
        \addplot[green, mark=square] table[x=Session, y={majority}]{tiered_imagenet_f1.dat};
        \addplot[blue!80, mark=*] table[x=Session, y={Agg. $f$-avg}]{tiered_imagenet_f1.dat};
        \coordinate (bot) at (rel axis cs:1,0);
    \end{groupplot}
    
\path (top|-current bounding box.north)--
      coordinate(legendpos)
      (bot|-current bounding box.north);
\matrix[
    matrix of nodes,
    anchor=south,
    draw,
    inner sep=0.2em,
  ]at([yshift=-90ex]legendpos)
  {
    \ref{plot:shallow}& Shallow NN &[5pt]
    \ref{plot:deep}& Deep NN &[5pt]
    \ref{plot:wa}& Weighted average NN &[5pt]
    \ref{plot:arithmetic}& Arithmetic &[5pt] \\
    \ref{plot:geometric}& Geometric &[5pt]
    \ref{plot:harmonic}& Harmonic &[5pt]
    \ref{plot:majority}& Majority vote &[5pt]
    \ref{plot:ours2}& AFA \\};
\end{tikzpicture}
\end{adjustbox}
	\caption{Comparison of ensemble learning output fusion methods, on FSCIL datasets, in terms of averaged F1 score over all classes, for various sessions $k$. The closer the F1 score is to 1, the better. As sessions are incrementally added, the classification task becomes increasingly more difficult because of the additional categories to be handled.} \label{plot:naive_avg}
\end{figure*}
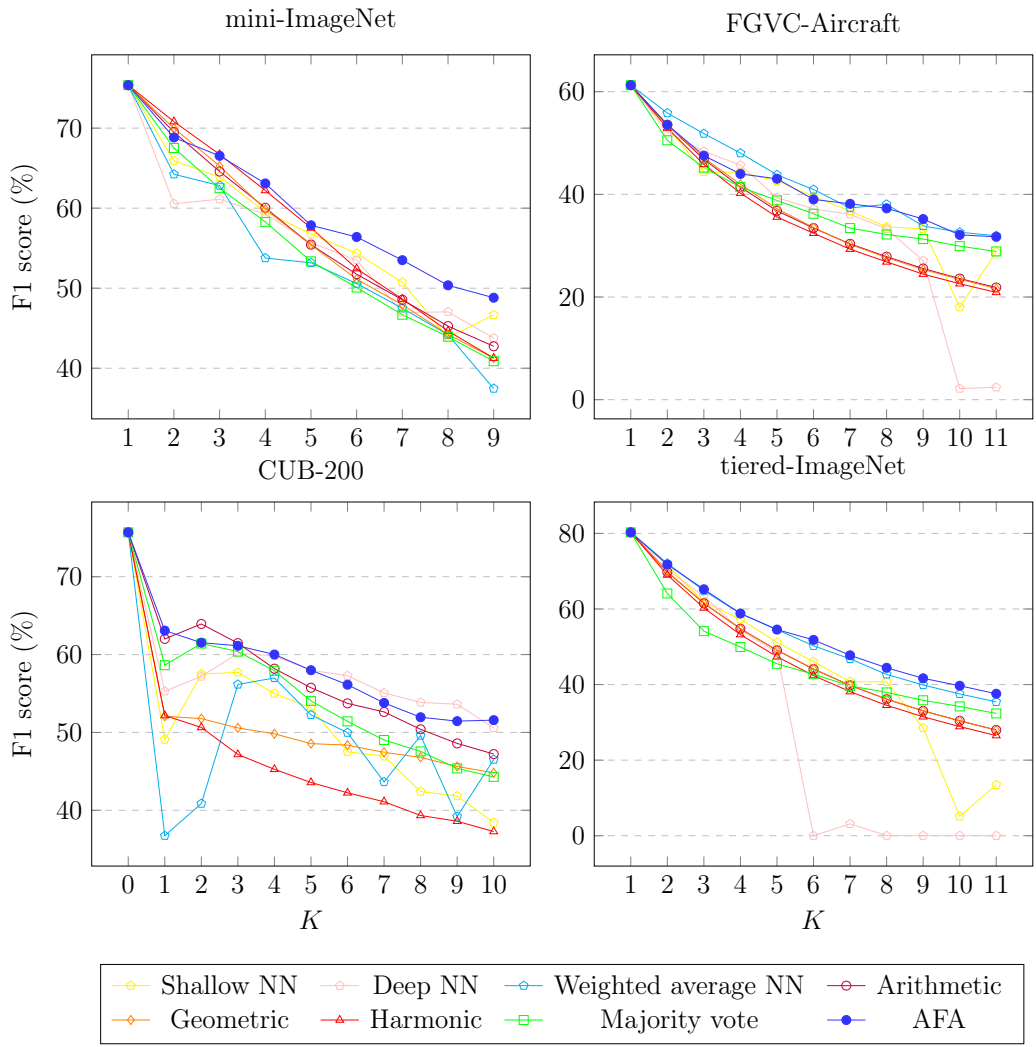

\paragraph{Comparison of AFA with state-of-the-art FSCIL approaches}

First, a global inspection of the results show that our proposed method AFA shows comparable results with respect to FSCIL-specific methods such as CEC \cite{zhang2021few} and FACT \cite{zhou2022forward}. The F1 score of our method is higher on mini-ImageNet and close to the state-of-the art on CUB-200, tiered-ImageNet and FGVC-Aircraft. This confirms the ability of our proposed ensemble learning pipeline to tackle the FSCIL challenging scenario. An advantage of our method is its simple design, with interpretable and easy to learn modules.

\paragraph{Comparison of AFA with standard average} We now focus on the comparison between AFA with classic ensembling strategies, based on simple averaging using either arithmetic, geometric, or harmonic mean. We also display results obtained with a majority vote. As Tables \ref{tab:f1_sota_min} and  \ref{tab:f1_sota_cub} (bottom part), and Fig.~\ref{plot:naive_avg} show, classic averaging methods show quite inconsistent results. The best type of average seems to depend on session number and dataset. Similarly, the majority vote scheme seems to produce worse results in early sessions, when few voters are available, but a gain in performance with an increasing number of voters. The AFA ensemble method we propose models and combines, by design, those different types of average. It learns their optimal balancing, thus producing significantly better performance in all sessions.

\paragraph{Comparison of AFA with standard neural networks} Performance of our proposed method were also compared against three different types of ensembling neural network, namely (1) a shallow neural network with similar number of parameters and layers as our AFA model, (2) a deeper neural network with five fully connected layers, and (3) a neural network specifically designed for ensembling including a weighted average layer followed by a fully connected layer for the output \cite{weightedavgnn}. All neural network models, including the AFA model, were trained with the same process with only slight adjustments on learning rate parameters to adapt to each model architecture, for the sake of fair comparisons. The results, in terms of accuracy and F1 scores (the larger, the better), are summarized in Table \ref{tab:f1_sota_min} and Figure \ref{plot:naive_avg}.

\begin{figure}[ht!]
    \centering
	\includegraphics[width=.95\textwidth]{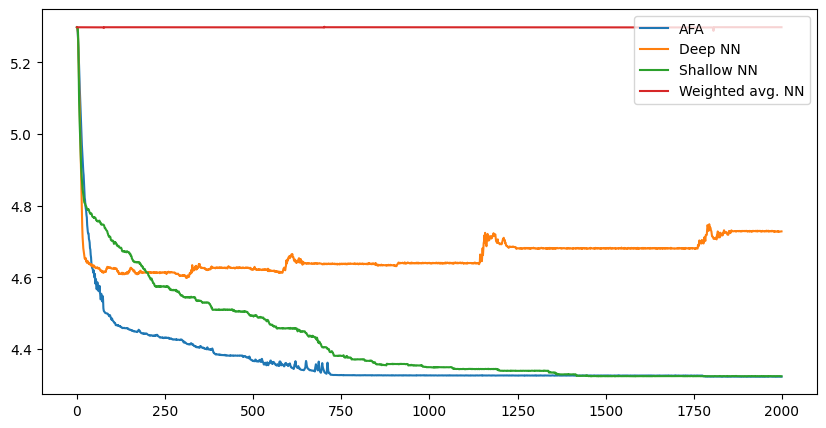}
	\caption{Loss versus epoch number for various NN-based ensemble learning methods, including our proposed AFA, on the last (10th) session of the CUB-200 setting.}
    \label{fig:loss}
\end{figure}

AFA outperforms those other NN approaches. The second best performing is the shallow NN, indicating that the limited number of parameters from both AFA and the shallow NN produces a decisive advantage when compared to the deep NN. The latter shows inconsistent results across different datasets. For example, it performs better in last session of mini-ImageNet than the weighted average NN (F1 score of 43.77\% against 37.46\%) but worse in last session of CUB-200 (38.43\% against 46.52\%). Indeed, these later sessions drastically increase its architecture size due to the higher number $K$ of models to ensemble and the higher number of classes $N_K$ to predict, while not providing a significantly larger training set because of the few-shot constraint, making the training of larger models difficult.

Figure \ref{fig:loss} illustrates these behaviours. It displays the training losses of the different stacked NN models, including the proposed AFA, on the last (10th) session for the CUB-200 setting. The plots demonstrate our model stability and fast convergence to a lower loss value than other methods which exhibit a plateau at a higher value. The deep NN loss behaviour shows the inability to train a large number of parameters on such limited amount of data, with a validation loss diverging. The shallow NN, having a similar number of parameters as our AFA model, also converges but with a lower rate and to a higher loss value, as demonstrated by its F1 score which is 1.01 \% lower than the AFA model. Finally, having only $k$ parameters, the Weighted Average NN weights converge in a single epoch as shown by its loss that does not evolve along subsequent epochs. 

\section{Conclusion}
In this work, we have proposed a novel approach for ensemble learning
that we have applied to the FSCIL problem. The FSCIL setting requires to produce methods able to handle an increasing number of classes with the extra constraint that only a few number of training samples are available for novel classes. FSCIL state-of-the-art extensively adopts meta-learning strategies, similarly to the few-shot classification setting. We propose here a conceptually simpler approach based on the ensembling of a set of nearest-neighbour classifiers. While basic ensembling strategies do not perform well in a consistent manner, our innovative Aggregated $f$-averages (AFA) ensembling model reaches significantly better and more stable performance than all its ensemble learning competitors, while achieving results comparable to the most sophisticated state-of-the-art FSCIL models. AFA is a supervised neural network with a specific architecture, specific activation functions, and specific weight constraints that models and combines various types of averages. AFA has a reduced number of trainable weights and present the additional advantage of being interpretable. In future work, we intend to further illustrate the flexibility and efficiency of our AFA model on regression tasks.


\bibliographystyle{elsarticle-num} 
\bibliography{bibliography}
\end{document}